\documentclass[conference]{IEEEtran}
\IEEEoverridecommandlockouts
\usepackage{cite}
\usepackage{amsmath,amssymb,amsfonts}
\usepackage{algorithmic}
\usepackage{textcomp}
\usepackage{xcolor}
\def\BibTeX{{\rm B\kern-.05em{\sc i\kern-.025em b}\kern-.08em T\kern-.1667em\lower.7ex\hbox{E}\kern-.125emX}}
    
\begin{document}

\title{
Intensional Artificial Intelligence:
From Symbol Emergence to Explainable and Empathetic AI
%\thanks{Note: Once accepted we are happy to pay the \$100 charge for an additional page as specified on the conference website.}
}

\author{\IEEEauthorblockN{Michael Timothy Bennett}
\IEEEauthorblockA{\textit{School of Computing} \\
\textit{The Australian National University}\\
Canberra, Australia \\
michael.bennett@anu.edu.au}
\and
\IEEEauthorblockN{Yoshihiro Maruyama}
\IEEEauthorblockA{\textit{School of Computing} \\
\textit{The Australian National University}\\
Canberra, Australia \\
yoshihiro.maruyama@anu.edu.au}
}

\maketitle

\begin{abstract}
We argue that an explainable artificial intelligence must possess a rationale for its decisions, be able to infer the purpose of observed behaviour, and be able to explain its decisions in the context of what its audience understands and intends. To address these issues we present four novel contributions. 
Firstly, we define an arbitrary task in terms of perceptual states, and discuss two extremes of a domain of possible solutions. 
Secondly, we define the intensional solution. Optimal by some definitions of intelligence, it describes the purpose of a task. An agent possessed of it has a rationale for its decisions in terms of that purpose, expressed in a perceptual symbol system grounded in hardware. 
Thirdly, to communicate that rationale requires natural language, a means of encoding and decoding perceptual states. We propose a theory of meaning in which, to acquire language, an agent should model the world a language describes rather than the language itself. If the utterances of humans are of predictive value to the agent's goals, then the agent will imbue those utterances with meaning in terms of its own goals and perceptual states. 
In the context of Peircean semiotics, a community of agents must share rough approximations of signs, referents and interpretants in order to communicate. 
Meaning exists only in the context of intent, so to communicate with humans an agent must have comparable experiences and goals. An agent that learns intensional solutions, compelled by objective functions somewhat analogous to human motivators such as hunger and pain, may be capable of explaining its rationale not just in terms of its own intent, but in terms of what its audience understands and intends. It forms some approximation of the perceptual states of humans.
Finally, we introduce the Mirror Symbol Hypothesis which posits that symbols learned as part of an intensional solution, by an agent labouring under human-like compulsions, behave like mirror neurons. These neurons, according to popular speculation, may play an important role in the facilitation of empathy.
\end{abstract}


\section{Introduction}
    \label{intro}
    %\subsection{Motivating Narrative}
    Artificial intelligence is a tool, and a tool is what we make of it. Perhaps one envisions a future of plenty in which the ever diminishing necessity of human labor, the automation of the mundane, gives way to a benevolent artificial superintelligence that solves all our problems. Liberated from the shackles of necessity we explore and indulge every whim.
    Yet when we think of the future we do not envision the utopian dreams of Ian M. Banks. A tool is only useful so long as it serves the intended purpose for which it is employed. Automation is not new, so why is automation using learning algorithms any different from the automation of the past 200 years? 
    
    We trust a hammer because it obeys constraints we understand. We trust cryptocurrency and imbue it with value because it obeys constraints at least some people understand. These things are reliable and understandable, the same standard we demand of safety and mission critical systems such as aircraft landing gear.
    Yet we cannot say the same of humans, so how can we trust ourselves? A human can explain its intent. A human has integrity if its stated intent is consistent with its behaviour and, if the human's intent is aligned with our own, we may trust it. We have a degree of certainty that future behaviour will fall within the bounds of what we consider acceptable. More, because one knows a human models one's own intent, interpreting what one means rather than just says, one trusts it to actively seek behaviour that aligns with one's desires. Humans empathise.
    
    Learning algorithms are tools which learn and change. We cannot trust black box learning machines because we cannot be certain of their behaviour before the fact nor explain, to our satisfaction, any rationale after the fact \cite{b1, b2, b3}. 
    We suggest that the reason black box learning models cannot explain their rationale is because there \textit{is} no rationale for their actions.  
    Consider GPT-3, the recent state of the art in natural language processing \cite{b4}. It can write convincing essays, perform basic arithmetic and even make passable jokes. Yet it clearly does not understand what it does. Given ``solve for x: x + 4 = 10" it returns ``6", but given ``solve for x: x + 40000 = 100000" it returns ``50000". It has not understood arithmetic and its rationale, if it can even be said to have one, is mere correlation.
    
    One might think causal modelling \cite{b5, b6} will solve the issue. Certainly a causal model will have what we might consider reasons for a prediction. However, we explain our own reasons for doing anything in terms of loosely defined high level objects, of what we consider most relevant to the listener, not raw sensorimotor stimuli and certainly not in exhaustive detail. A causal model applied to a task of sufficient complexity would produce a rationale so complex it may as well be a black box. Indeed, there are many tasks of such complexity that we ourselves cannot comprehensively identify or agree upon the constraints that determine what is desirable, even in terms of high level objects. For example, an ethical artificial intelligence (AI) would need to learn and understand social and moral norms, assuming we don't wish for a live re-enactment of Asimov's work \cite{b7, b8, b9}. How can we construct a machine capable of inferring the constraints under which we wish it to operate, which we ourselves don't understand? That can communicate those constraints in a manner we do understand? 
    
    This suggests a machine that understands what is meant rather than said using natural language. In a programming language meaning is exact, specified down to the smallest detail; but natural language is an emergent phenomenon \cite{b10} in which meaning hinges upon context and loose interpretation. State of the art natural language models identify correlations in syntax well enough, yet just as they do not understand they do not mean what they say. What is meaning, and how is it computed? Such questions are of broad practical significance even to the likes of Deepmind \cite{b11}. 
    
    If such a complex agent is to function alongside humans it must learn social and moral norms. It must infer what humans feel and desire. In other words, it must be able to empathise.
    
    To answer these questions we draw upon ideas from such disparate fields as semiotics, biology and algorithmic information theory, to posit a possible solution from first principles.
    
    Section \ref{symbolemergence} summarises background material. Section \ref{rules} defines the arbitrary task and intensional solutions. In section \ref{explainable} we introduce a theory of meaning that explains how natural language can arise within a lower level perceptual symbol system. Section \ref{empathy} introduces the Mirror Symbol Hypothesis, how meaning may give rise to an agent capable of empathy. 
    
	\section{Symbol Emergence}
	\label{symbolemergence}
    That the mind and body would be separable seems only natural, after all we construct hardware and software as separate entities. Such inclinations gave rise to the notion of a ``Physical Symbol System" \cite{b12}, in which symbols with assumed intrinsic meaning facilitate abstract reasoning, cognition. However, a top down symbol system must somehow connect with hardware, with low level sensorimotor stimuli. This is the ``Symbol Grounding Problem" \cite{b13}, in which Harnad asked (to paraphrase) ``how is a symbol connected to that which it signifies?" After all, the utterance ``cat" is meaningless without knowing anything of the object it represents. He deemed the pursuit of artificial cognition from a top-down perspective ``parasitic on the meanings in the head of the interpreter".
    
    This is not to say that cognition takes place absent any symbolic manipulation, but Harnad's work seems to suggest we ask how the body might give rise to the mind, rather than be connected to one. Working from first principles up to an emergent high level symbol system we explain cognition as embodied, situated and extending into an environment. In other words, enactive cognition \cite{b14}.
    
    What then is the basic unit of cognition? Prior to the 20$^{\text{th}}$ century cognition was primarily conceived of in terms of perceptual states, and so Barsalou \cite{b15} proposed Perceptual Symbol Systems. The basic unit of cognition is then a perceptual symbol, a subset of the perceptual state, the state of hardware. For example the perceptual symbol system of a computer is the arrangement of transistors which physically implement memory and logic. 
    
    How does a high level symbol come into being? First, let us examine the notion of a high level symbol. One might be tempted to define a symbol as dyadic, as consisting only of a sign and a referent to which that sign connects. However, such a picture is incomplete. Meaning in natural language is inconsistent, ``parasitic on the meanings in the head of the interpreter" as Harnad put it. Hence triadic Peircean semiosis seems more appropriate, in which the sign and referent are connected by an interpretant, which determines their effect upon the interpreter.
    
    This is the foundation upon which field of symbol emergence stands, constructing agents that observe and interact with their environment to learn the meaning of words \cite{b16}. The agent observes the co-incidence of multimodal sensorimotor stimuli and generates its own symbol system \cite{b10}. Of similar design is DALL-E, a version of GPT-3 trained on multimodal text and image data \cite{b17}, can visualise what a word means and so in some limited sense learns a high level symbol from low level sensorimotor data. However it is still based upon GPT-3, which mimics rather than understands arithmetic. Something is missing.
	
	\section{Intensional and Extensional Solutions}
	\label{rules}

	Assume for a moment we have two agents, both equally capable of solving a task. One of them is a mimic who understands nothing of the purpose they serve, while the other acts with intent.
	Capability at a task held constant, to an outside observer there is no perceivable difference between these two. However, there is an observable difference in the rate at which they learn.
	
	A mimic learns the decisions necessary to serve a purpose, but not the purpose itself. An intentional agent may learn those decisions but it must, by definition, first and foremost know the purpose of a task. A mimic does not know that the goal has been satisfied, only that they have reproduced an action previously deemed successful. Conversely, an intentional agent can evaluate any given course of action to determine whether it will satisfy the goal. It learns what is intended by an action, rather than just the action itself. Subsequently while a mimic may need to observe a large number of examples illustrating what is considered correct to master a task, the intentional agent may master a task having observed only enough examples to infer the task's purpose. Given the same number of training examples, it will choose correct decisions at least as frequently as the mimic and, depending on the task, generalise to situations as yet unencountered. 
	
	To examine how an agent capable of inferring what is intended might be constructed, we'll now formalise these ideas in terms of perceptual symbol systems and states.

	\subsection{Continuous Variables are Unnecessary}

    To do so it is necessary to clarify the notion of a perceptual state. Though it is suggested perceptual symbol systems may involve continuous variables \cite{b15}, we argue that only discrete variables need be considered. First, representing continuous values with arbitrary precision implies the ability to store an unlimited amount of information. It seems reasonable to assume that physical limitations upon memory exist, and so representing the domain of a continuous variable may be impossible. Second, even if a continuous domain could be represented, sensors capable of recording an infinite amount of detail do not to our knowledge exist, making continuous representation rather pointless. Third, actuators capable of infinite precision would be required to take advantage of such precise sensors. As a result, either the capacity to store information, the capacity to collect information or the number of decisions which can be made must be finite, and so for practical purposes we are justified in treating all variables as discrete.
    
	\subsection{A General-Purpose Notion of Tasks}
    
    To learn purpose, and by extension general concepts or definitions, it is helpful to define an arbitrary task. Such an all encompassing notion might be conceived of as follows;
	
	\begin{itemize}
	    \item A finite set $\mathcal{X} = \{X_1, X_2, ..., X_n\}$ of binary variables.
	    \item A set $\mathcal{Z}$ of every complete or partial assignment of values of the variables in $\mathcal{X}$, where
	    \begin{itemize}
	        \item an element $z \in \mathcal{Z}$ is an assignment of binary values $z_k$, which is $0$ or $1$, to some of the variables above, which we regard as a sequence $\langle X_i = z_i, X_j = z_j, ..., X_m = z_m \rangle$, representing a perceptual state.
	    \end{itemize}
	    \item A set of goal states $\mathcal{G} = \{z \in \mathcal{Z} : C(z)\}$, where
	    \begin{itemize}
	        \item $C(z)$ means that $z$ satisfies, to some acceptable degree or with some acceptable probability, some arbitrary notion of a goal.
	    \end{itemize}
	    \item A set of states $\mathcal{S} = \{s \in \mathcal{Z} : V(s)\}$ of initial states in which a decision takes place, where
	    \begin{itemize}
	        \item $V(s)$ means that there exists $g \in \mathcal{G}$ such that $s$ is a subsequence of $g$, in other words for each state in $\mathcal{S}$ there exists an acceptable, goal satisfying supersequence in $\mathcal{G}$.
	    \end{itemize}
	\end{itemize}
    The process by which a decision is evaluated is as follows:
    \begin{enumerate}
        \item The agent is in state $a \in \mathcal{S}$.
        \item The agent selects a state $b \in \mathcal{Z}$ such that $a$ is a subsequence of $b$ and writes it to memory.
        \item If $b \in \mathcal{G}$, then the agent will have succeeded at the task to an acceptable degree or with some acceptable probability.
    \end{enumerate}
    Note that this makes no comment about the state following $b$, of which $b$ may not be a subsequence.
    
	To abduct $b \in \mathcal{G}$ from a given $a \in \mathcal{S}$ requires a constraint sufficient to determine whether any partial assignment $b$ is in $\mathcal{G}$. This is similar to the notion of a ``unified theory that explains the sequence" \cite{b18} from recent artificial general intelligence research, however we differ significantly from Evans et al. in what constitutes the desired explanatory constraint, along with assumptions about time sequences, predicates, objects and so forth. Given our premises, notions like ``object" must emerge from the enaction of cognition.
	
	This explanatory constraint can be written (using a perceptual symbol system) as a sentence which is true of a given $s \in \mathcal{Z}$ if and only if $s \in \mathcal{G}$. As not all states in $\mathcal{G}$ assign a value to all variables, a form of three valued logic (``true", ``false" and ``indeterminate") may be useful (though perhaps not strictly necessary). Many such explanatory sentences may exist, but there is a helpful notion of intensional and extensional definitions to be found in the philosophy of language \cite{b19}. For example, the extensional definition of the game Chess is the enumeration of every possible game of Chess, while the intensional definition could be the rules of chess. Thus we'll consider two extremes:
	\begin{enumerate}
	    \item The extensional definition of the task. This is a sentence $D$ enumerating every member of $\mathcal{G}$ as a long disjunction ``a or b or c or ... ". It may also be thought of as the category of all goal satisfying states.
	    \item The intensional definition of the task. This is a sentence $C$ stating only what we might naturally consider the concept or rules of the task.
	\end{enumerate}
	The rules $C$ of a game intuitively tend to be formed from the weakest, most general statements necessary to verify whether any given example of a game is legal, and sufficient to abduct every possible legal game. Conversely, the category $D$ enumerating all valid games is formed from the strongest, most specific statements possible. It is important to note that, given $\mathcal{S}$, $C$ is necessary and sufficient information to reconstruct $D$. Either of ($C$ and $\mathcal{S}$) or $D$ is necessary and sufficient to reconstruct $\mathcal{G}$. 
	
	\subsection{The Weakest Solution}
	\label{weakest}
	While the notion of intensional and extensional definitions is not new, the notion of intensional and extensional solutions as two extremes of a domain of possible solutions to an arbitrary task, is \cite{b20}.
	
	An agent that models $C$ is attempting to infer the goal or purpose of a task. An agent that models $D$ is attempting to master the task through mimicry.  
	Both the concept $C$ and the category $D$ would be sufficient knowledge for an agent to perform optimally. However, learning typically relies upon an ostensive \cite{b21} definition; a small set of examples serving to illustrate how a task works, or what kind of states satisfy the goal. Such an ostensive definition could be constructed as follows;
	\begin{itemize}
	    \item A set $\mathcal{G}_o \subset \mathcal{G}$ of goal satisfying states, which does not contain a supersequence of every member of $\mathcal{S}$.
	    \item A set $\mathcal{S}_o = \{s \in \mathcal{S} : B(s)\}$ of initial states in which a decision takes place, where
	    \begin{itemize}
	        \item $B(s)$ means that there exists $g \in \mathcal{G}_o$ such that $s$ is a subsequence of $g$.
	    \end{itemize}
	\end{itemize}
	From this ostensive definition an agent could derive $D_o$, a category of goal satisfying states, or $C_o$, the concept or rules of the task necessary and sufficient to reconstruct the ostensive definition. 
	An agent who learns $D_o$ cannot generalise to $D$, they will eventually encounter a state $s \in \mathcal{S}$ where $s \not\in \mathcal{S}_o$ for which it has no solution. On the other hand, if the ostensive definition is sufficiently representative of the task, then $C_o$ necessary and sufficient to imply $\mathcal{G}_o$ given $\mathcal{S}_o$ will also be necessary and sufficient to imply $\mathcal{G}$ given $\mathcal{S}$, meaning the agent will have mastered the task using only the ostensive definition (in which case $C_o = C$).
	
	The weaker the statements from which a solution is formed (whilst remaining both necessary and sufficient to imply $\mathcal{G}_o$ given $\mathcal{S}_o$), the larger the subset of $\mathcal{S}$ can be for which the sentence implies goal satisfying states. For brevity we'll say solution $a$ is weaker than a solution $b$ if it is formed from weaker statements.  
	
	In this sense $C_o$ is the weakest possible solution necessary and sufficient to imply $G_o$ given $S_o$, and $D_o$ is the strongest. We argue that an agent that models the weakest necessary and sufficient constraint would therefore learn at least as fast (in terms of number of examples required) as an agent modelling any stronger constraint. In the worst case scenario (a perfectly random, uniform distribution of goal satisfying states with no observable cause and effect relationships to model), the weakest and strongest constraints would be the same (enumerating all goal satisfying states), and the entire contents of $\mathcal{G}$ would need to be observed in order to master the task. In the best case scenarios (simple, consistent cause and effect relationship between aspects of the initial perceptual states and what constitutes a goal state), there would be a stark difference between the weakest and strongest constraint as the weakest would state that cause and effect relationship and nothing else. If as Chollet \cite{b22} suggests, intelligence is the ability to generalise, then these two constraints represent the product of its extremes.  
	
	$C$ is not only the most general solution, but also represents intent. Intent is the desire for the outcome of an action to possess a particular quality \cite{b23}. Intent is usually communicated by describing that desired quality, rather than the every possible outcome possessed of that quality. For example, no one would describe the intent to realise victory in a game of Chess by stating in excruciating detail every possible game of Chess for which this aspiration is realised. Instead, the intent is better described by a small number of the weakest possible statements necessary and sufficient to determine, given any game of Chess, whether that intent is satisfied. Yes, it is arguable that an agent may ``intend" to mimic a set of actions, but stating in general terms the qualities actions ought to possess, and by extension what they ought to achieve, seems a more natural definition of the word. Thus we consider an agent for which $C$ is a goal to possess the intent $C$.
	
	Just as lossless compression exploits causal relations \cite{b24}, any decision based upon the weakest possible constraint sufficient to verify membership of $\mathcal{G}_o$ must discount any event not strictly necessary to maximise the certainty of that decision. An event in this case is any aspect of a partial assignment of values. This is akin to conditional independence in a probabilistic graphical model, whereby a variable may be eliminated from consideration given the available information provided by other variables. To constrain possible solutions using any part of an assignment of values which does not further minimise uncertainty given other events have been observed would create a stronger constraint than necessary, and so violate the definition of $C$. However, the constraint must remain both necessary and sufficient, taking note of exceptions to any rule, or occasional ``Black Swan Events" \cite{b25}. As a result, the agent modelling $C$ will identify cause and effect relationships and be equipped to entertain counterfactuals \cite{b5, b6}.
	
	We will name $C$ the \textbf{intensional solution} of a task. It represents what the highest possible ``G-factor" \cite{b22} would produce, the ability to generalise from any sufficient ostensive definition to solve every instance of a task, and importantly \textit{why}. If a constraint describes what a number of states share in common sufficient to reconstruct all of those states, then it has found information sufficient to losslessly compress those states. This is similar to the idea that intelligence is the ability to compress information expressed by Chaitin \cite{b26}, Hutter \cite{b27}, Legg \cite{b28} and others. Hutter also defines superintelligence or universal artificial intelligence as the ability to perform ``optimally across a wide range of environments". The scope of task being arbitrary, an intensional agent situated in any one of a wide range of environments, equipped as AIXI \cite{b27} is with a reward function, could construct ostensive definitions of optimal behaviour and then find $C$. As $C$ is the weakest possible constraint, it would (with as small a sufficient ostensive definition as possible) generalise from the ostensive definition of optimal behaviour to perform optimally across the environment as a whole. As such an agent that generates intensional solutions meets Hutter's definition of superintelligence (albeit limited by its sensorimotor system). An example of how such a reinforcement learning agent might pursue intensional solutions is given in \cite{b29}. Unlike Solomonoff's universal prior, the intensional solution appears to be computable. At the time of writing, a prototype has been implemented which finds the intensional solution to a variety of string prediction problems, solving them as expected. Further investigation is required, however results should be forthcoming in 2021.
	
	That said, the \textbf{extensional solution} $D$ is not without merit as the abduction of correct responses would require minimal computation. It is akin to a lookup table, a polynomial fit to past observations, those solutions a human might intuit without being able to identify the why of it all. The ideal balance may be to seek both, as the intensional solution would facilitate generalisation while the extensional solution act as a heuristic to efficiently guide the abduction of goal satisfying states.
	
	\section{From Intensional to Explainable AI}
	\label{explainable}
	It is abundantly obvious that to explain the ``why" of an action, an agent must possess a reason for having undertaken it. Just as ``everybody else was doing it" is not considered a satisfactory rationale for the behaviour of a child, mimicry is not a satisfying rationale for the behaviour of an AI. This suggests that both intention, and the identification of cause and effect relationships, are necessary conditions of explainability. 
	
	To empathise, an agent must infer what is meant rather than said, and communicate in kind. An explainable AI that mimics a plausible explanation without knowing what it means is still just a mimic. While existing theories of meaning are commendable \cite{b30}, in the context of intensional solutions and symbol emergence, we present what we believe is a more compelling alternative:
	\begin{enumerate}
	    \item To learn semantics, the agent should model the world a language describes, rather than the language itself. We seem to imbue symbols with meaning given their significance to our goals (driven by motivators such as hunger and pain). In other words, meaning only really exists given intent. 
	    \item While specification of intent in natural language would require semantics, we consider intent to be an intensional solution pursued as a goal. It is written in the physical state of hardware, constructed in a perceptual symbol system that is in no way ``parasitic on the meanings in the head of the interpreter". 
	    \item An agent situated within a community of humans may observe utterances in natural language. If those utterances are of no significance in predicting aspects of the world relevant to its goals, then they will be ignored. However, if utterances are of predictive value, then an agent that constructs intensional solutions will identify those utterances as patterns within its perceptual states useful in predicting what to do next. It imbues them with meaning, in the context of its own goals and experience. If context or intent changes, then so does meaning. 
	\end{enumerate}
    Not just spoken words but any phenomena may be imbued with meaning. For example, an agent compelled by hunger may recognise the sound of a dinner bell indicates the likely availability of food at a specific location. It identifies cause and effect because it pursues intensional solutions. It imbues the sound of the dinner bell with meaning, making the dinner bell a symbol in terms of its own experience and compulsion to eat. Language is then the means by which agents encode, transmit and decode perceptual states inasmuch as they pertain to goals. Meaning is the implications with respect to a goal.
	
	To explore how meaningful natural language may emerge from the enaction of cognition, we begin by illustrating how an unnatural language, machine code, may be acquired through observation. 
	
	\subsection{Illustrative Example: Instruction Set Architecture}
	The task ``performing the role of a CPU" could be defined as follows:
	\begin{itemize}
	    \item The set $\mathcal{X}$ of variables representing all bits in memory, in the program counter and other registers, in every component of the computer the moment before an operation, along with a duplicates representing the moment after an operation has taken place.
	    \item The set $\mathcal{Z}$ of complete or partial assignments of binary values to the variables in $\mathcal{X}$.
	    \item The set $\mathcal{G} = \{z \in \mathcal{Z} : C(z)\}$ of goal satisfying states representing correct execution of an instruction.
	    \item The set $\mathcal{S} = \{z \in \mathcal{Z} : V(z)\}$ of initial states for which there exists $g \in \mathcal{G}$ of which $z$ is a subsequence. 
	\end{itemize}
	The extensional solution $D$ must, for each and every legitimate combination of OPCODE and OPERAND, describe an initial subsequence in which the program counter points at the instruction, and a goal satisfying supersequence in which the instruction has been correctly executed. Consider the operation ADD which takes as input the values contained in two registers, adds them together, and then deposits the result in a destination register. The extensional solution must enumerate every possible combination of destination and input registers, as well as every possible legal combination of input and output values. Conversely, the intensional solution would stipulate only the logic necessary to validate binary addition and correctly interpret register labels. Such logic is conventionally written in the physical arrangement of transistors, from which the correct interpretation of every combination of OPCODE and OPERAND may be deduced. The intensional solution differs from the CPU itself in that it stipulates what must be true of goal states, rather than actually enacting those changes. An intensional solution to binary addition has been constructed by the prototype mentioned in \ref{weakest}, and behaves as anticipated.
	
	\subsection{Semantics of Natural Language}
	\label{semantics}
	The intensional solution to an instruction set architecture hopefully now seems obvious. The perceptual symbol system is the physical arrangement of transistors and the instruction set architecture is specified using the perceptual symbol system. In the same manner any higher level language may be constructed, one upon the other as assembly is built upon machine code, compiled programming language upon assembly, and interpreted programming language upon compiled in a conventional computer. 
	
	Machine code details a sequence of explicit instructions. There is one and only one correct interpretation of an instruction, although the form may vary. These interpretations were hard coded by a human in the arrangement of transistors, rather than being what we might recognize as emergent phenomena. 
	
	In contrast natural language is an emergent phenomenon, brought about by the interaction of agents labouring under similar compulsions. Though among humans compulsions may be almost universal, the experiences associated with a sign and thus meaning in terms of those compulsions vary from human to human. However, it is clear that loosely shared interpretants, signs and referents are sufficient for communication, as we are not only able to communicate effectively with one another but with other social animals such as dogs.
	
	Consider an agent compelled by pain, hunger and other motivators under which biological organisms labour. In any given a state these objective functions have an expected value, which can be used to define a partial ordering of states. This describes a task, the goal of which is to transition into those states in which the expected values of objective functions are maximised. That goal would change as what the agent considers optimal depends upon past experience. By interacting with the environment, the agent constructs sets of states representing optimal behaviour, ostensive definitions which in turn are used to hypothesise intensional solutions of improving quality. This describes a kind of non-parametric reinforcement learning agent.
	
	If the agent is situated within a community of humans that share information using natural language, then that information (if observed) will be part of its perceptual state. It need not be compelled to learn the language. An intensional solution relies only upon that information absolutely necessary and sufficient to reproduce the ostensive definition from which it was derived. If the behaviour of the humans is of no value in predicting goal satisfying states then it will be ignored. However, if patterns (subsequences of states) in the human's behaviour are useful in predicting which states will satisfy the agent's goal, then the intensional solution may describe those patterns and identify any causal relationships. It relies upon signals transmitted by humans to model and predict the world, and so is learning semantics; what the signals mean inasmuch as they are relevant to the agent's goal. 
	To interpret a signal, the agent transitions from a state in which the signal is observed to one which recalls past experience associated with that signal. To transmit a signal, the agent transitions from a state in which it behooves the agent to affect change in the perceptual a state of a human, into one in which its actuators communicate the relevant signs. By learning $C$ that recalls the agent's own experiences of phenomena associated with a symbol, with the utterances and actions of humans, the agent is modelling human perceptual states in terms of its own past experience. Such an agent models human intent. Because $C$ is a constraint, a sentence in a perceptual symbol system, that part-of-sentence which is informs correct transmission of a signal is the same as that which interprets that symbol. 
	
	\section{From Explainable to Empathetic AI}
	\label{empathy}
	To properly model human perceptual states and communicate with us on our own terms, an agent must have experiences analogous to our own. Intensional solutions may imbue words such as ``pain" and ``hunger" with some meaning, iterations of which may better inform objective functions designed to approximate those experiences, the human condition. However, empathy requires an agent not only feel, but feel some approximation of what another individual experiences as the agent observes them.
	
	In the human brain mirror neurons are suspected to play an important role in the facilitation of empathy \cite{b31}. These neurons trigger both when an individual performs an action, and also when that individual observes another performing that same action \cite{b32}. For example, the observation of an individual laughing makes the observer more inclined to laugh. We feel something of what we observe others doing, in other words we empathise.
	
	Empathy allows us to model the perceptual states of others. Much like mirror neurons, a symbol in natural language learned as part of an intensional solution would be a true part-of-statement both when the agent observes an action, and when that same agent performs the action. Recall, $C$ plays the role of interpretant, transitioning the agent from a perceptual state in which they observe a sign associated with a symbol, into one in which they recall their own experience associated with that same symbol. Likewise if the agent wishes to communicate its own experiences, it can transmit signs associated with those symbols closest to what it means to convey.
	
	We call this The Mirror Symbol Hypothesis, the notion aspects of an intensional solution mimic the behaviour of mirror neurons in the human brain, assuming the speculation regarding their existence and significance is correct. In any case, compelled by objective functions analogous to human feeling the agent may grasp something of the perceptual state of those humans it interacts with, and in so doing empathise.
	
	Empathy would allow the agent to better communicate with humans, as its choice of which signs to transmit would be informed by the estimation of perceptual states and by the presumed intentions of the audience, rather than just the conditional probability of those signs as with other methods. As such it may constitute a significant boon in the ability to explain a decision. Not only might the agent explain its actions in terms of what it intended in broad, intensional terms, but by knowing something of how its audience perceives the world it may communicate in terms that specific audience can understand.
	
	\section{Concluding Remarks}
	\label{conclusion} 
	We've illustrated three key requirements for explainable AI. First, the agent must possess a rationale for its decisions beyond mere correlation, which implies causal reasoning and intent. Second, the agent must learn the purpose of a set of decisions, rather than merely which decisions are correct. Third, it must be able to summarise and explain an otherwise incomprehensibly complex rationale in a manner that its audience can understand.
	
	We've defined an arbitrary task in terms of perceptual states, and proposed the intensional and extensional solutions defined by their relative weakness, learned from an ostensive definition of a task. An agent that learns intensional solutions will know why a decision was taken, what purpose it served, and so possess a rationale for its decisions. In contrast the extensional solution describes goal satisfying decisions, yet does not represent purpose or intent, cannot generalise and cannot explain those decisions. Further, as an intensional solution can be learned from a smaller ostensive definition than any stronger constraint, an agent that seeks it will learn at least as fast as an agent that seeks any stronger constraint. In other words, it is the optimal means of learning any arbitrary task.
	
	Abducting correct decisions from an intensional solution by brute force search may be intractable. Something close to the extensional solution, perhaps a function fit to previously observed goal satisfying states, may compliment the intensional solution as a heuristic. A similar idea is evident in the design of AlphaGo \cite{b33}, which may be described in greatly simplified terms as being constructed of a dynamic component that learns and a static component which does not. What was learned amounted to a heuristic, employed by the static component informed by the rules of Go and the state of the board to generate sequences of moves. The rules of Go alone don't make it obvious which sequence of moves is most likely to result in a win. The task they describe is ``how to play Go", while the heuristic was constructed for the task ``decide which sequence of moves is best". The static component employed the intensional solution to one task while the heuristic more closely resembled the extensional solution to another. Because of this, AlphaGo possessed a rationale for deciding whether a sequence of moves was legal, but no rationale beyond correlation for deciding whether such a sequence is ``good", which may be why the developers struggled to explain why it made the decisions it did.
	
	A task being of arbitrary scope, we may define one of such breadth and complexity that natural language becomes necessary. Utterances of value in satisfying a goal will be imbued with meaning. As this meaning is learned as part of the intensional solution to a task, the agent will communicate its rationale for decisions in terms of semantics and intent, rather than correlations in syntax. 
	
	This new theory of meaning is of philosophical as well as technical significance. While the emphasis on mental state and intent bears some similarity to the Gricean foundational theory of meaning \cite{b30}, the notion of an intensional solution as intent represents a significant advancement. Put simply, it posits meaning and language exist to improve an organism's ability to predict its environment and co-operate with other organisms. Behaviour is imbued with meaning as part of an intensional solution, a byproduct of the compulsions under which an organism labours.
	
	Finally, to infer the purpose of human behaviour an agent must have experience comparable to the human condition. Labouring under compulsions analogous to those of a human an agent may glean something of what a human intends and understands, and so communicate complex concepts in comprehensible manner. As symbols learned as part of an intensional solution may behave as mirror neurons, it is arguable that such an agent may empathise. 
	
	This raises interesting possibilities for hybrid approaches to artificial intelligence such as Neuralink, as piggybacking on the nervous system of a human may be more achievable than developing objective functions that accurately reflect the human experience. While such lofty ideas may be beyond the scope of immediate practical concern, they serve to illustrate where intensional solutions and the ability to learn semantics may eventually lead.


\begin{thebibliography}{00}
        \bibitem{b1} Wojciech Samek, Thomas Wiegand, and Klaus-Robert Muller. Explainable Artificial Intelligence: Understanding, Visualizing and Interpreting Deep Learning  Models. 2017. arXiv: 1708.08296[cs.AI]
        \bibitem{b2} Amina Adadi and Mohammed Berrada. “Peeking Inside the Black-Box: A  Survey on Explainable Artificial Intelligence (XAI)”.  In: IEEE Access 6 (2018), pp. 52138–52160. %DOI: 10.1109/ACCESS.2018.2870052.
        \bibitem{b3} Alejandro Barredo Arrieta et al. “Explainable Artificial Intelligence (XAI): Concepts, taxonomies, opportunities and challenges toward responsible AI”. In: Information Fusion 58 (2020), pp. 82–115. %ISSN: 1566-2535. DOI: https://doi.org/10.1016/j.inffus.2019.12.012. URL: http://www.sciencedirect.com/science/article/pii/S1566253519308103.
        \bibitem{b4} Luciano Floridi and Massimo Chiriatti. “GPT-3: Its Nature, Scope, Limits, and Consequences”. In: Mindsand Machines(2020), pp. 1–14.
        \bibitem{b5} Judea Pearl. Causality: Models, Reasoning and Inference. 2nd. USA: Cambridge University Press, 2009. ISBN: 052189560X.
        \bibitem{b6} Judea Pearl and Dana Mackenzie. The Book of Why: The New Science of Cause and Effect. 1st. USA: BasicBooks, Inc., 2018. %ISBN: 046509760X.
        \bibitem{b7} Michael Timothy Bennett and Yoshihiro Maruyama. Philosophical Specification of Empathetic Ethical Artificial Intelligence. Manuscript. 2021.
        \bibitem{b8} Matthias Scheutz.   “The Case for Explicit Ethical Agents”. In: AI Magazine 38.4 (2017), pp. 57–64. %DOI:10.1609/aimag.v38i4.2746. URL:  https://ojs.aaai.org/index.php/aimagazine/article/view/2746.
        \bibitem{b9} Isaac Asimov. Runaround. 1942.
        \bibitem{b10} Tadahiro Taniguchi et al. “Symbol emergence in robotics: a survey”. In: Advanced Robotics (2016), pp. 706–728. %URL: https://doi.org/10.1080/01691864.2016.1164622.
        \bibitem{b11} Adam Santoro et al. Symbolic Behaviour in  Artificial Intelligence. 2021. arXiv: 2102.03406 [cs.AI].
        \bibitem{b12} Allen Newell. “Physical symbol systems”. In: Cog. Sci(1980), pp. 135–183.
        \bibitem{b13} Stevan Harnad. The Symbol Grounding Problem. 1990.
        \bibitem{b14} Evan Thompson. Mind in Life. Biology, Phenomenology and the Sciences of Mind. Vol. 18. Jan. 2007.
        \bibitem{b15} Lawrence W. Barsalou. “Perceptual symbol systems”. In: Behavioral and Brain Sciences 22.4 (1999), pp. 577–660.
        \bibitem{b16} Tadahiro Taniguchi et al. “Symbol Emergence in Cognitive Developmental Systems: A Survey”. In: IEEE Transactions on Cognitive and Developmental Systems 11.4 (2019), pp. 494–516.
        \bibitem{b17} Aditya Ramesh et al. DALL-E: Creating Images From Text. https://openai.com/blog/dall-e/. 2021.
        \bibitem{b18} Richard Evans et al. “Making sense of sensory input”. In: Artificial Intelligence 293 (2021), 103438. %ISSN:0004-3702. DOI:  https://doi.org/10.1016/j.artint.2020.103438. URL:  http://www.sciencedirect.com/science/article/pii/S0004370220301855.
        \bibitem{b19} Gary Ostertag. “Emily Elizabeth Constance Jones”. In: The Stanford Encyclopedia of Philosophy. Ed. by Edward N. Zalta. Fall 2020. Metaphysics Research Lab, Stanford University, 2020.
        \bibitem{b20} Michael Timothy Bennett. PhD Thesis Manuscript. Australian National University. 2021.
        \bibitem{b21} Anil Gupta. “Definitions”. In:The Stanford Encyclopedia of  Philosophy. Ed. by Edward N. Zalta. Winter2019. Metaphysics Research Lab, Stanford University, 2019.
        \bibitem{b22} Francois Chollet. On the Measure of Intelligence. 2019. arXiv: 1911.01547[cs.AI].
        \bibitem{b23} Kieran Setiya. “Intention”. In:The Stanford Encyclopedia of Philosophy. Ed. by Edward N. Zalta. Fall 2018. Metaphysics Research Lab, Stanford University, 2018.
        \bibitem{b24} Kailash Budhathoki and Jilles Vreeken. “Origo: causal inference by compression”.  In: Knowledge and  Information Systems56.2 (2018), pp. 285–307. %URL:  https://doi.org/10.1007/s10115-017-1130-5.
        \bibitem{b25} Nassim Nicholas Taleb. The Black Swan : The Impact of The Highly Improbable. First edition. New York: Random House, 2007.
        \bibitem{b26} Gregory Chaitin. “The limits of reason.” In: Scientific American 294 3 (2006), pp. 74–81.
        \bibitem{b27} Marcus Hutter. Universal artificial intelligence: Sequential decisions based on algorithmic probability. Springer, 2005.
        \bibitem{b28} Shane Legg. “Machine Super Intelligence”. 2008.
        \bibitem{b29} Michael Timothy Bennett and Yoshihiro Maruyama. Conceptual Foundations of Intensional AI: A Computational Theory of Meaning. Manuscript. 2021.
        \bibitem{b30} Jeff Speaks. “Theories of Meaning”. In: The  Stanford Encyclopedia of Philosophy. Ed. by Edward N.  Zalta. Winter 2019. Metaphysics Research Lab, Stanford University, 2019.
        \bibitem{b31} Giacomo Rizzolatti, Maddalena Fabbri-Destro, and Marzio Gerbella. “The Mirror Neuron Mechanism”. In: Reference Module in Neuroscience and Biobehavioral Psychology. Elsevier, 2019. %URL: http://www.sciencedirect.com/science/article/pii/B9780128093245235712.
        \bibitem{b32} Giacomo Rizzolatti. “Action Understanding”. In: Brain Mapping.   Ed. by Arthur W. Toga. Waltham:   Academic Press, 2015, pp. 677–682. %URL: http://www.sciencedirect.com/science/article/pii/B9780123970251003547.
        \bibitem{b33} David Silver et al. “Mastering the game of Go  with deep neural networks and tree search”. In: Nature 529 (Jan. 2016), pp. 484–489. %DOI: 10.1038/nature16961.
    \end{thebibliography}
\end{document}